RAISE: A self-driving laboratory for interfacial property formulation discovery

Mohammad Nazeri [1†], Sheldon Mei [2†], Jeffrey Watchorn [3†], Alex Zhang [4], Erin Ng [2], Tao Wen [4], Abhijoy Mandal [5], Kevin Golovin [4], Alán Aspuru-Guzik [3,5,6,7], Frank Gu [1,2,3*]

[1] Institute of Biomedical Engineering, University of Toronto, 164 College St, Toronto, ON M5S 3G9

[2] Department of Chemical Engineering and Applied Chemistry, University of Toronto, 200 College St, Toronto, ON M5S 3E5

[3] Acceleration Consortium, 700 University Avenue, Toronto, ON M5G 1X6

[4] Department of Mechanical & Industrial Engineering, University of Toronto, 5 King's College Rd, Toronto, ON M5S 3G8

[5] Department of Computer Science, University of Toronto, Toronto, Ontario, M5S 3H6, Canada

[6] Department of Chemistry, University of Toronto, Toronto, Ontario, M5S 3H6, Canada

[7] Vector Institute for Artificial Intelligence, 661 University Ave. Suite 710, ON M5G 1M1, 15 Toronto, Canada

Corresponding author's email address
f.gu@utoronto.ca

Graphical Abstract

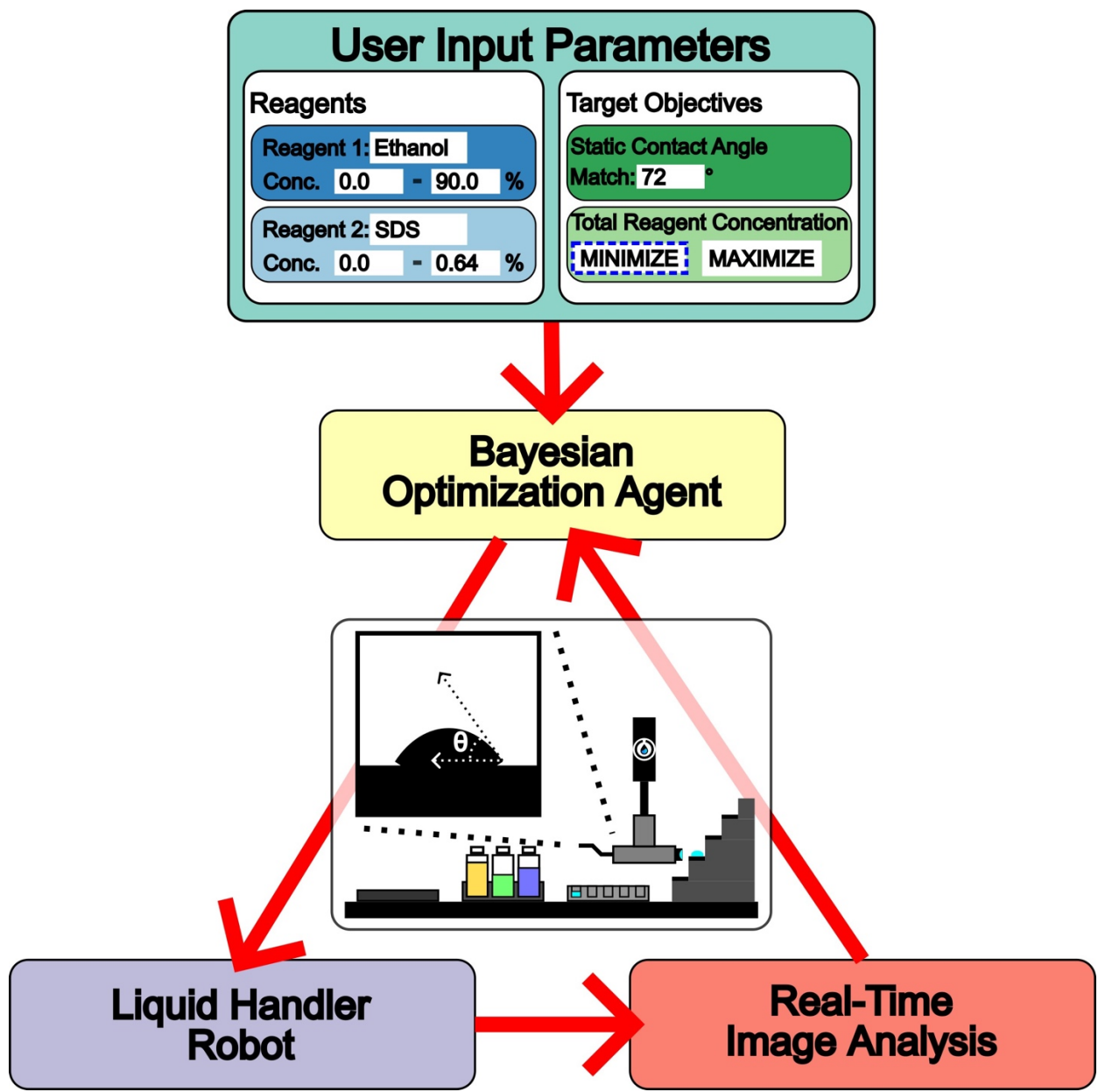

**Abstract**

Surface wettability is a critical design parameter for biomedical devices, coatings, and textiles. Contact angle measurements quantify liquid-surface interactions, which depend strongly on liquid formulation. Herein, we present the Robotic Autonomous Imaging Surface Evaluator (RAISE), a closed-loop, self-driving laboratory that is capable of linking liquid formulation optimization with surface wettability assessment. RAISE comprises a full experimental orchestrator with the ability of mixing liquid ingredients to create varying formulation cocktails, transferring droplets of prepared formulations to a high-throughput stage, and using a pick-and-place camera tool for automated droplet image capture. The system also includes an automated image processing pipeline to measure contact angles. This closed loop experiment orchestrator is integrated with a Bayesian Optimization (BO) client, which enables iterative exploration of new formulations based on previous contact angle measurements to meet user-defined objectives. The system operates in a high-throughput manner and can achieve a measurement rate of approximately 1 contact angle measurement per minute. Here we demonstrate RAISE can be used to explore surfactant wettability and how surfactant combinations create tunable formulations that compensate for purity-related variations. Furthermore, multi-objective BO demonstrates how precise and optimal formulations can be reached based on application-specific goals. The optimization is guided by a desirability score, which prioritizes formulations that are within target contact angle ranges, minimize surfactant usage and reduce cost. This work demonstrates the capabilities of RAISE to autonomously link liquid formulations to contact angle measurements in a closed-loop system, using multi-objective BO to efficiently identify optimal formulations aligned with researcher-defined criteria.

# 1. Introduction

Surface wettability plays a critical role across diverse applications, from textiles [1,2] and food packaging [2–4] to biomedical devices [5–7] and agricultural sprays [8–10]. While the surface itself plays a key role in determining wettability [11,12], liquid formulation optimization for target surface wettability remains a fundamental challenge across many industries from biomedical devices [13] to advanced coatings [14], where formulation composition and interfacial properties strongly affect wetting behavior [15–17]. While contact angle measurement provide a widely used and precise means of quantifying wettability [18–20], the measured angle is influenced by formulation-dependent liquid properties such as presence of surfactants [21,22], which change the balance of interfacial tensions at the solid-liquid-vapor boundary [23,24]. Current approaches, however, treat formulation development and surface characterization as discrete and disconnected processes [25,26]. This separation prevents systematic exploration of high-dimensional formulation spaces where physical properties, performance, cost, and regulatory constraints need to be simultaneously considered [27].

The existing formulation discovery workflows typically involve trial and error experimentation guided by experience and intuition [28,29], rather than data-driven optimization [30]. Manual measurements such as contact angle, though accurate, operate as stand-alone characterization tools that cannot efficiently navigate the complex, nonlinear interactions between multiple chemical components. The resulting workflow are time-intensive and unable to systemically balance target objectives. On the other hand, static contact angle, which is the most adopted metric in industry due to its simplicity, is highly sensitive to several experimental parameters that can affect measurement accuracy [31,32]. These include the droplet dispensing mechanism and the volume of liquid droplet, especially on non-ideal substrates. For example, in terms of droplet deposition on the substrate, studies have shown that applying droplets by lowering a syringe needle, moving the stage, or manually pipetting can each give different static contact angles on the same sample [33]. Additionally, manual measurements require a skilled operator and significant time and remain susceptible to variability from factors such as contamination, evaporation, and ambient conditions, which limits reproducibility across studies [34,35].

While automation efforts have addressed individual components, such as MULA as a liquid handling platform [36], formulation robots such as Polybot [37], and other semi-automated contact angle system, there is a lack of platform that can integrate formulation preparation, surface characterization, and multi-objective optimization within a single autonomous workflow. Several semi-automated systems have been developed that focus on individual components of the workflow, such as image processing [38–41], syringe control mechanisms, tilting stages for dynamic contact angle measurements [42], and automated liquid dispensing [43–45]. While these systems have improved consistency and reduced operator error in specific tasks, they still fall short of delivering a fully autonomous solution. This fragmentation can be a limiting factor to enable training and predictions of complex formulation where optimization may require hundreds or thousands of experiments.

To address these limitations, we introduce RAISE, Robotic Autonomous Imaging Surface Evaluator, a closed-loop self-driving platform that links liquid formulation optimization with contact angle measurement through Bayesian optimization. RAISE combines automated formulation preparation, high throughput surface characterization using custom hardware and an integrated image processing pipeline, and multi-objective optimization to systemically navigate formulation spaces while balancing performance, cost, and other constraints related to materials usage. By enabling closed-loop discovery, this system can be classified as a self-driving laboratory [46].

**2. Materials and Method**

2.1. Materials

The high-quality reagents were purchased from Sigma-Aldrich, and included sodium dodecyl sulfate (SDS) 99% (CAS # 151-21-3), Tween 20 (CAS # 9005-64-5), Titrin X-100 (polyethylene glycol tert-octylphenyl ether) (CAS # 9036-19-5), Decyl glucoside (n-Decyl β-D-glucopyranoside, DGO) (CAS # 58846-77-8), and ethanol (CAS # 64-17-5). Commercial-grade reagents included sodium dodecyl sulfate (SDS) 94% (CAS # 151-21-3) and Tween 20 (CAS # 9005-64-5), both purchased from EZ Chemicals. All chemical reagents were mixed with Milli-Q water (MQ) unless otherwise stated.

2.2. Hardware Development

2.2.1. Camera Holders

As part of this work, we developed two different camera adapters to enable camera imaging within an Opentrons OT-2 platform that mount with two different methods: the press-fit method or the pick-and-place tool (Figure 1 (a))

**Press-fit Camera Holder**

The press-fit camera holder (Figure 1 (a)) is fabricated using 3D-printed poly(lactic acid) (PLA) resin. It is designed in the shape of a pipette to securely press-fit onto the pipette. The holder features a long vertical attachment and a horizontal neck that provides stable support for a horizontally mounted camera. While this method is beneficial in some ways, particularly when we need to prevent shaking of the gantry, since the camera is mounted directly on the stable part of the pipette gantry, it also limits pipetting capabilities because it occupies one of the pipettes, leaving only one available for use. Additionally, having the camera and holder fixed to one of the pipettes introduces a collision risk with labware on the deck during movements.

**Pick-and-place Camera Holder**

The pick-and-place camera holder offers a flexible solution that preserves full pipette availability during operation. As show in our previous work [47], the pick-and-place mechanism involves inserting an end-cut pipette tip into the custom 3D-printed camera holder module. The OT-2 pipette gantry then treats the module like a pipette tip, which enables it to transfer the camera to different zones on the deck. As shown in Figure 1 (a), this system consists of two main components: the camera-to-pipette adapter that the pipette can grip, and a base holder that remains fixed on the OT-2 deck. The camera adapter features a cylindrical cavity designed to securely hold the horizontally oriented camera, mounted on a short horizontal neck to enable alignment between the camera aperture and the sample imaging stage. The base holder is designed with angled walls and a V-shaped notch that guide and stabilize the adapter when it is returned to the base holder. These features restrict rotation: the fit with the pipette tip mechanically constrains rotation about the X and Y axes, while the V-shaped notch in the base holder restricts rotation about the Z axis. Together, these designs ensure that the camera remains securely aligned and properly positioned during movements across the deck. During operation, the pipette can autonomously engage with or release the camera adapter without interfering with other tasks, allowing for the integration of real-time imaging. Compared to the press-fit method, the pick-and-place system minimizes collision risks and maintains full functionality of both pipettes. Additionally, for both camera holder designs, the horizontal cylindrical camera includes designated screw holes to securely affix the camera to the holder.

2.2.2. Stage and Lighting Systems

**Wrapped LED Multilevel Stage**

The initial experimental setup consisted of a multilevel stage structure designed to support contact angle measurement experiments. This stage featured horizontal stepped levels, allowing for flexible positioning and sample mounting. A critical design element was the integration of a wrapped LED light source that encircled the stage levels as shown in Figure 1 (c). This configuration provided consistent illumination around the sample, enhancing the visibility of droplet profiles on different surfaces. However, the wrapped LED system had limitations in terms of brightness, leading to suboptimal image contrast. The resulting images often exhibited reduced clarity, making it difficult to accurately discern the droplet profile for reliable contact angle measurements. The uneven and dim illumination hindered the effectiveness of image processing techniques. Additionally, the physical complexity of maintaining the wrapped LED structure limited operational stability.

**Flat-Panel LED Multilevel Stage**

To address these issues, a redesigned experimental stage was developed. This version incorporated a more streamlined multilevel platform with horizontal levels, optimized for sample stability and droplet application. A key enhancement was the replacement of the wrapped LED system with a flat-panel LED positioned directly behind the stage. The flat-panel LED configuration shown in Figure 1 (d), provided significantly brighter and more uniform backlighting, resulting in enhanced image contrast. This modification minimized glare and shadowing effects, which previously affected measurement accuracy. The homogeneous and high-intensity lighting facilitated clearer droplet visualization, supporting more reliable image processing and contributing to higher measurement precision.

2.2.3. Hardware Integration into Workflow

In addition to developing hardware modules, these components were integrated into an automated experimental pipeline (Figure *1* (b)). The camera mounts, lighting stages, and formulation tools were arranged to help execution of the four core experimental steps: formulation preparation, droplet deposition, droplet imaging, and contact angle measurement. Figure *1* (c) shows the complete deck layout, where the formulation zone, multilevel stage with backlighting, pick-and-place camera mount, and environmental sensor (temperature and humidity display) are positioned. This hardware integration supports reliable autonomous operation while preserving full pipetting functionality.

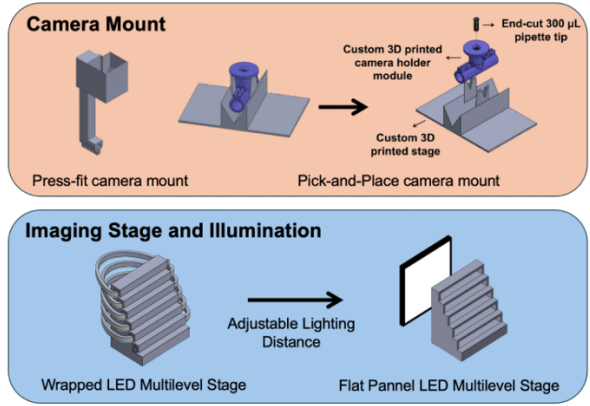
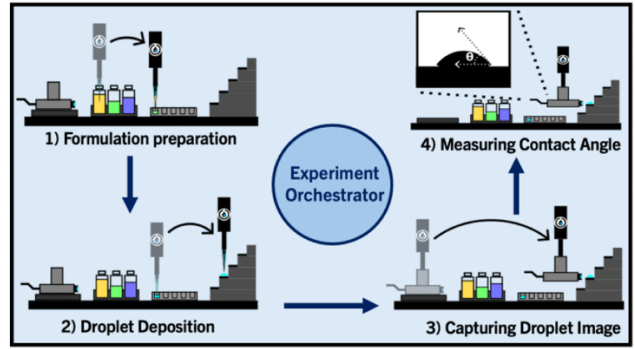
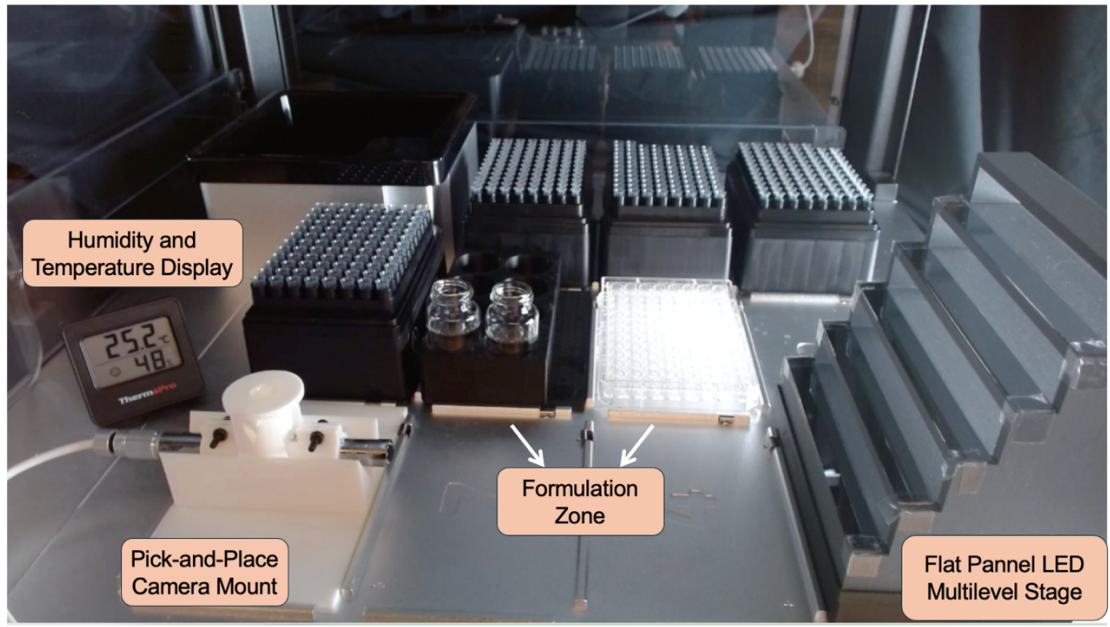

Figure 1. a) Hardware designs for the Opentron OT-2 platform. Press-fit camera mount attached directly to the pipette, pick-and-place camera mount consisting of a removable adapter and base holder, multilevel stage with wrapped LED illumination, and multilevel stage with flat-panel LED illumination for uniform backlighting. b) Automated workflow steps: formulation preparation, droplet deposition, droplet imaging, and contact angle measurement. (c) Deck layout with integrated hardware modules.

## 2.3. Software Development

### 2.3.1. Orchestrator & Internet of Things (IoT)

The RAISE platform is controlled by a Python-based orchestrator server that coordinates communication with the OT-2 over a local area network. The orchestrator launches a server that opens socket communication capabilities, and the OT-2 connects to the orchestrator server with a script uploaded onto the OT-2's on-board processor. A small light-weight camera connects to the orchestrator via USB to provide the RAISE platform image capturing capabilities. When experiments begin, the orchestrator first reads from several JSON configuration files that contains information for initial setup, including the predetermined layout of labware on the OT-2 deck, as well as the identities of liquids placed on the deck along with their corresponding stock concentrations. Afterwards, the orchestrator opens the camera and starts a static contact angle measuring process and a Bayesian Optimization (BO) client process. The orchestrator and BO client exchange information via a message queuing telemetry transport (MQTT) message broker. The BO client recommends new formulations for testing and the orchestrator returns measured contact angle values after completing each experiment. For each received formulation, the orchestrator performs a series of operations to conduct each experiment. First, the orchestrator calculates the working volumes of each liquid (µL) to mix, determines appropriate pipette tip usage for transferring each liquid, allocates an unused space on the deck for mixing, and allocates imaging locations on the RAISE stage. Secondly, the orchestrator instructs the OT-2 to perform the sequence of steps for mixing the sample formulation, placing small volumes liquid droplets onto the RAISE stage, picking up the camera, and imaging the droplet. Once the images are captured, they are automatically analyzed by the contact angle measuring process. This workflow enables fully autonomous experimental cycles without manual intervention between formulation preparation and data connection.

Figure 2 2.3.2. Bayesian Optimization (BO)

The RAISE platform integrates BayBe, a high-level BO library developed by Merck Group[48], which connects to the experimental orchestrator via a MQTT message broker (Figure 2). The BO client defines which chemical reagents will be tested and the concentration ranges for each reagent. The search space that the BO client explores is the set of unique concentration combinations of each reagent. This interaction loop is shown in Figure 2, linking formulation recommendations (top) with backend execution (bottom).

Each BO campaign is designed to target a specific contact angle. In each round of experiments, the BO client recommends a mixture of liquid reagents for the orchestrator to test. After the RAISE system performs the experiment and returns a static contact angle value, it sends the value back to the BO client via the message broker. Bayesian optimization builds surrogate models that capture the relationship between formulation parameters and the resulting contact angle. After a new experiment is completed, the contact angle measurement is normalized, and the normalized value becomes the latest datapoint incorporated into the surrogate model. The first formulation mixture that the BO client recommends is randomly selected, then it switches to a Bayesian recommender for future experiments. This surrogate model is based on a Gaussian Process [49], which defines a distribution over functions with mean function $m(x)$ and covariance function $k(x, x')$, where $x$ represents the formulation parameters (Equation (1)).

$$f(x) \sim gP(m(x), k(x, x')) \qquad (1)$$

After the posterior distribution is updated, an acquisition function is used to recommend the next experiment based on a policy that balances exploration (testing formulations with high uncertainty of the predictive model) and exploitation (testing formulations likely to yield contact angles close to the target). In our implementation, we used Thompson sampling [50] as the acquisition function in which a single function is sampled from the Gaussian Process posterior and the candidate that maximizes this sampled function is chosen for evaluation. This acquisition function favors exploration of uncertain regions of the search space.

A random seed is set prior to starting experiment campaigns to ensure that the order for experiment recommendations is reproducible. The BO client is programmed to run two types of campaigns: 1) a single-objective campaign, and 2) a multi-objective campaign. The single-objective campaign is designed to only seek out surfactant formulations that match a specified contact angle, while the dual-objected campaign is additionally designed to minimize total surfactant content in the mixture. Contact angle measurements and total surfactant measurements are normalized and averaged to assess the desirability of both target parameters together.

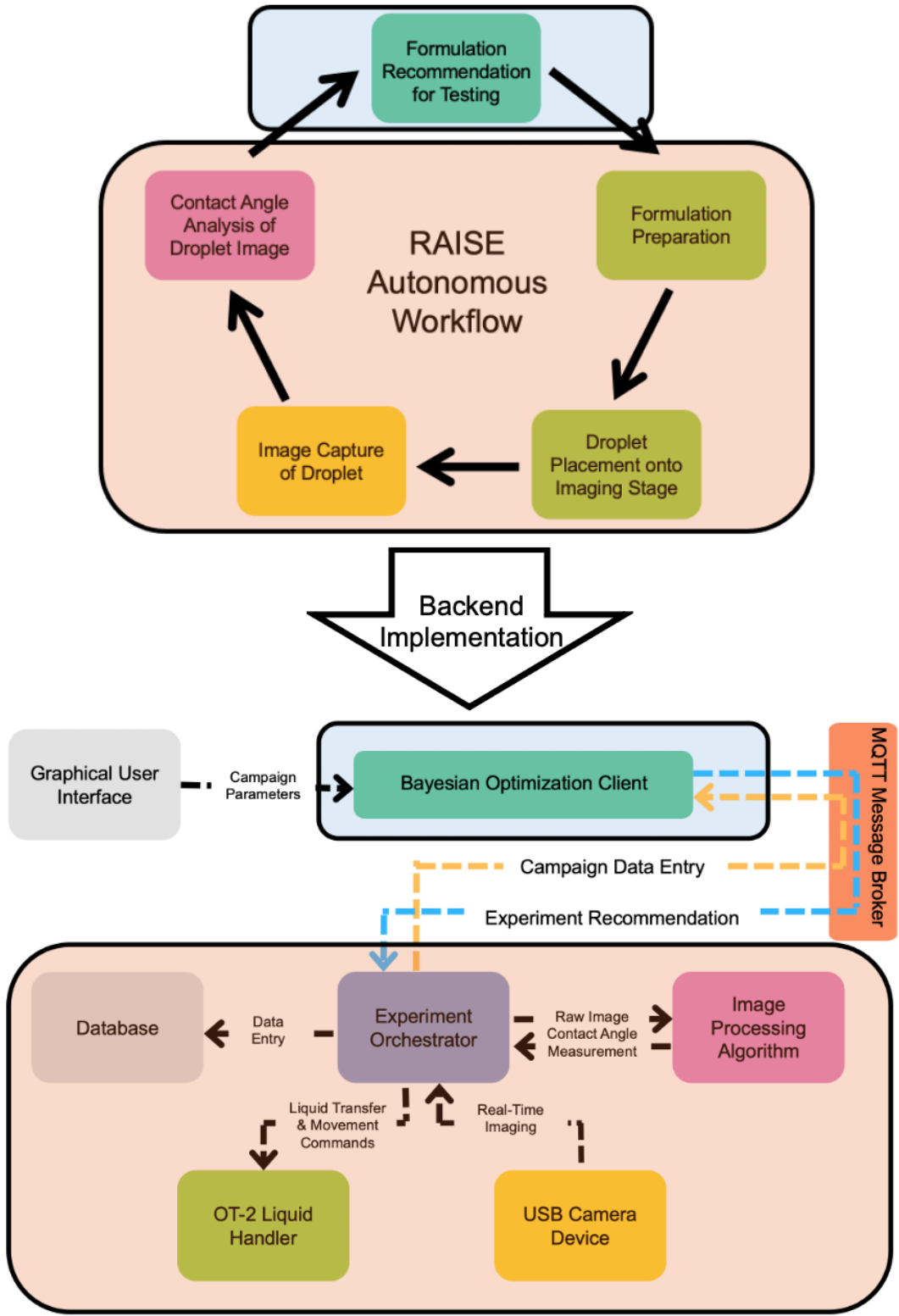

Figure 2. Architecture of the RAISE platform, showing the autonomous workflow (top) and backend implementation (bottom) integrating the orchestrator, hardware, image processing, and Bayesian Optimization client.

The RAISE system also has a web interface (shown in Figure 3) that allows users to enter inputs to design their own RAISE campaigns. Users can enter the names of liquids loaded on the deck of the OT-2 and their corresponding concentrations. Users can also input parameters to design BO campaigns that drive the close-loop system, including the ability to program the BO client to perform single- or multiple-objective campaigns. BO clients can be programmed to match a specified contact angle and either minimize or maximize total surfactant usage within a liquid formulation. After the user submits their inputs, they are saved in a format compatible with the orchestrator to run self-driving campaigns.

Figure 3. RAISE system web interface for setting up reagents and Bayesian optimization parameters for autonomous formulation campaigns.

2.4. Image Processing Pipeline for Static Contact Angle Measurement

A custom image analysis module written in Python has been developed to measure static contact angles of liquid droplets. Python-based libraries including scikit-image, Numpy, and matplotlib were used for image processing, manipulation, and displaying. Our image processing algorithm follows similar procedures as previously released open-source automated contact angle measuring algorithms, including Conan-ML and DropPy [51,52]. Though differing in implementation, each image processing algorithm first applies an image pre-processing procedure, which includes removing color from the image, cropping the image to isolate the droplet, and smoothing out noise. Afterwards, the processed images undergo edge detection to locate the boundary between the bright and dark regions in the image. Finally, a curve-fitting method is applied to the edges corresponding to the droplet above the surface and the static contact angle is calculated based on the fitted-curve. We wrote own custom image processing algorithm because the implementation of the other contact angle measuring algorithms was not compatible with our autonomous RAISE platform. Because images taken by the RAISE system show reflections beneath the droplets, the Conan-ML and DropPy algorithms encounter issues identifying the edges belonging to the droplets, resulting in poor curve-fitting and inaccurate static contact angle measurements.

To allow the RAISE system to autonomously and reliably measure static contact angle values from the images taken by the camera, we implanted our own image processing procedure that accounts for reflections beneath liquid droplets and properly identifies the edges belonging to the droplets. When the orchestrator captures an image, it is automatically saved and passed along to the image processing algorithm to begin measuring the contact angle. Images are analyzed concurrently while the OT-2 loads subsequent liquid samples onto the stage. The image processing procedure first includes converting the raw images to 8-bit grayscale images and then applying a Gaussian filter for remove noise. Next, a Sobel filter is applied to highlight the edges of the droplet on the grayscale image. A small square region surrounding the droplet is cropped from the grayscale image and resized to 1000 x 1000px. The resizing step helps increase accuracy of the contact angle measurement. Next, Otsu thresholding is performed on the resized image, which produces a binary image of the droplet on the surface. The contour between the bright and dark regions of the binary image are extracted and partitioned to isolate the contours belonging to the liquid-air interface above the surface. Extracted contours include regions belonging to a horizontal baseline, reflection of the droplet beneath the surface, and the actual droplet above the surface. The left and right contact points, associated with the surface-liquid-air interface, can be placed on the contour above the horizontal baseline based on the number of times the x-coordinate of each contour point changes direction moving from the top of the droplet to the edges of the image. The portion of the contour between the left and right contact points belong to the liquid-air interface of the droplet. After the contour of the liquid-air interface is identified, it undergoes fitting with the Bashforth-Adams equations to yield the final static contact angle measurement. We choose fitting with the Bashforth-Adams equations because it is the most rigorous method for modeling the Young-Laplace equation and yields the most accurate static contact angle measurements [52]. While both the Conan-ML and DropPy algorithms implement methods for Bashforth-Adams fitting [51,52], the fitting method by Conan-ML was incorporated into our final contact angle measuring algorithm. If the left and right contact points are not level, the image and contour are corrected by applying a slight rotation before curve-fitting ensure the most accurate fitting and contact angle measurement. This tilt correction is performed in the Conan-ML algorithm as well [51]. The root mean square error (RMSE) between fitted curves and detected droplet contours quantifies the quality of the curve-fitting. Figure 4 shows validation of the measurement, using automated detection of surfaces exhibiting static contact angles of 104° with water and 32° with ethanol. The code that comprises the RAISE system, including the image processing algorithm, has been uploaded to GitHub (https://github.com/AC-Formulations-SDL/RAISE).

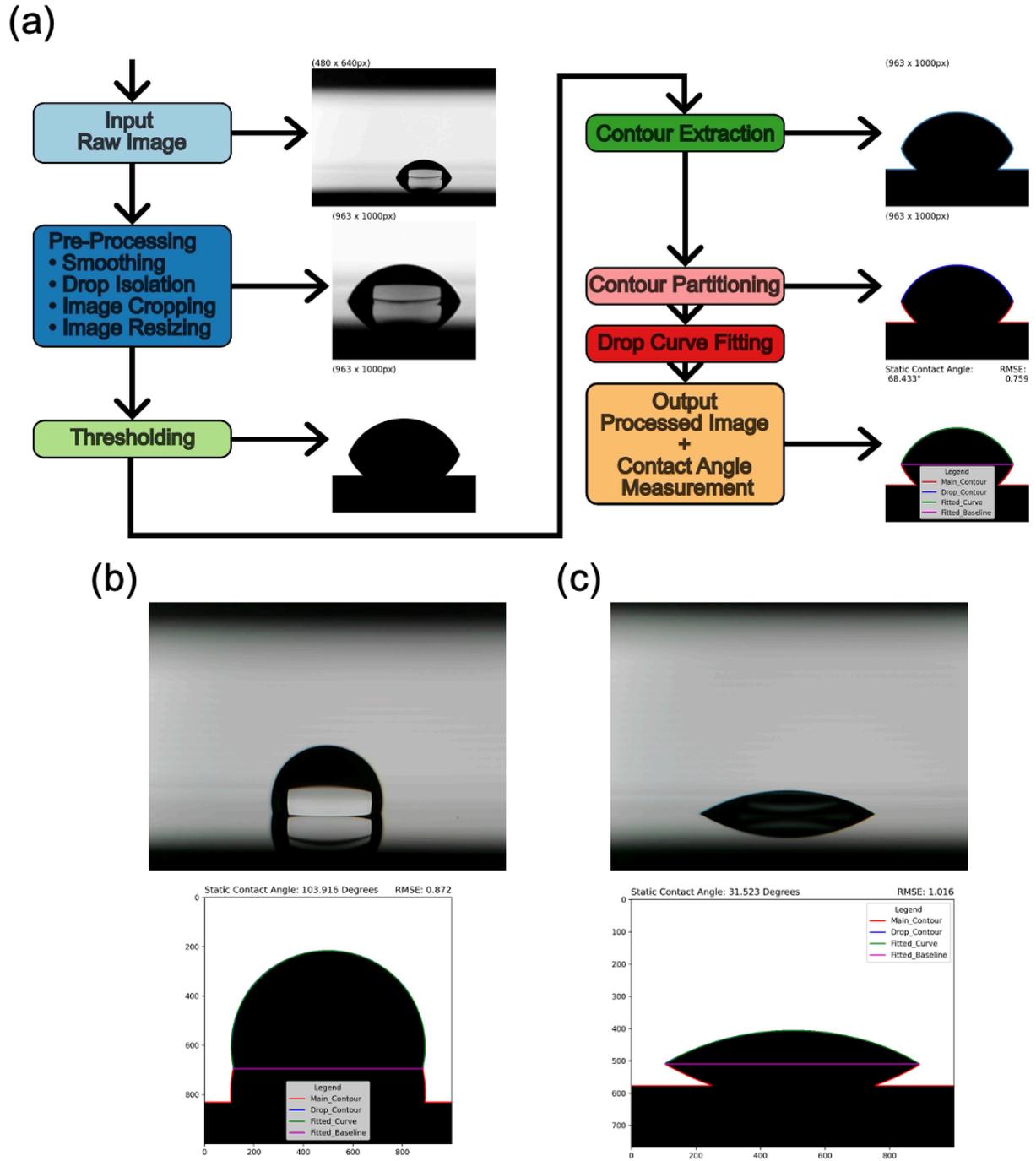

Figure 4Figure 4

Figure 4. Automated static contact angle measurements. (a) Image processing pipeline for droplet analysis, including pre-processing, thresholding, contour extraction, and curve fitting. (b) Water

droplet (0% ethanol) exhibiting a contact angle of ~104°. (c) Ethanol droplet (99.6%) exhibiting a contact angle of ~32°. RMSE values indicate the accuracy of the fitted droplet profile. All droplets were automatically prepared, deposited, imaged, and analyzed.

2.5. Manual Fabrication Protocols for Poly(dimethylsiloxane) (PDMS), Poly(tetrafluoroethylene) (PTFE), and Polystyrene (PS) Substrates

For PDMS substrates, single side polished silicon wafers with a thickness of 650 um and diameter of 150 mm were purchased from University Wafer Inc. Toluene, isopropyl alcohol (IPA), and 1,3-dichlorotetramethyldisiloxane were purchased from Gelest Inc. First, silicon wafers were cut into 120 × 10 mm substrates using a diamond scribing pen. Next, the silicon wafers were rinsed by toluene and isopropanol alcohol for 30 s using a squeeze bottle followed by compressed air drying. The dried silicon wafers were then treated with oxygen plasma at high power for 5 min at 106 mTorr vacuum pressure using a PDC-001-HP model from HARRICK PLASMA. After oxygen plasma, the silicon wafers were immediately transferred into 150 × 15 mm petri dishes with two small glass containers each holding 100 μL of 1,3-dichlorotetramethyldisiloxane. The petri dish was closed for 8 min, after which the coated silicon wafers were taken out from the petri dish and were rinsed by toluene and isopropyl alcohol for 1 min using squeeze bottles, followed by compressed air drying. The coating process was done in a glove box with controlled relative humidity of 20%. PTFE substrates (poly(tetrafluoroethylene); McMaster-Carr) were obtained commercially and were used as received after rinsing with IPA and drying with compressed air. PS substrates (polystyrene; Fisher Scientific) were also purchased commercially; before use, they were rinsed with IPA and then dried with compressed air.

## 3. Result and Discussion

3.1. Benchmarking Performance and Accuracy

RAISE accuracy was benchmarked against a conventional Ramé-Hart 260 Contact Angle Goniometer using ~3μL water droplets on PDMS, PTFE and PS substrates. This volume was chosen to minimize gravitational effects while maintain measurement stability on flat surfaces [33]. Manual goniometer measurement included 5 replicates per substrates, while RAISE performed 15 replicates per substrate. The data collected by RAISE showed strong agreement with conventional goniometry across all substrates (Figure 5). RAISE consistently showed improved precision with standard deviation reduction of 20.8% on PDMS, 29.1% on PS and 61.9% on PTFE samples. The improvement for PTFE was statistically significant (p-value=0.0106), likely due to eliminated operator variability in droplet placement and optimized contract line detection. In addition, the two stages on the RAISE platform processed approximately 30 formulations (n=3) per 90-minute cycle. This can be projected to about ~600 droplet measurement per 10-hour day shift, compared to ~60 samples for manual goniometry limited by operator handling, factoring in cleaning processes after each sample measurement. The 10X throughout improvement, combined with the reduced variability, shows the potential for RAISE to function for high throughput formulation screening.

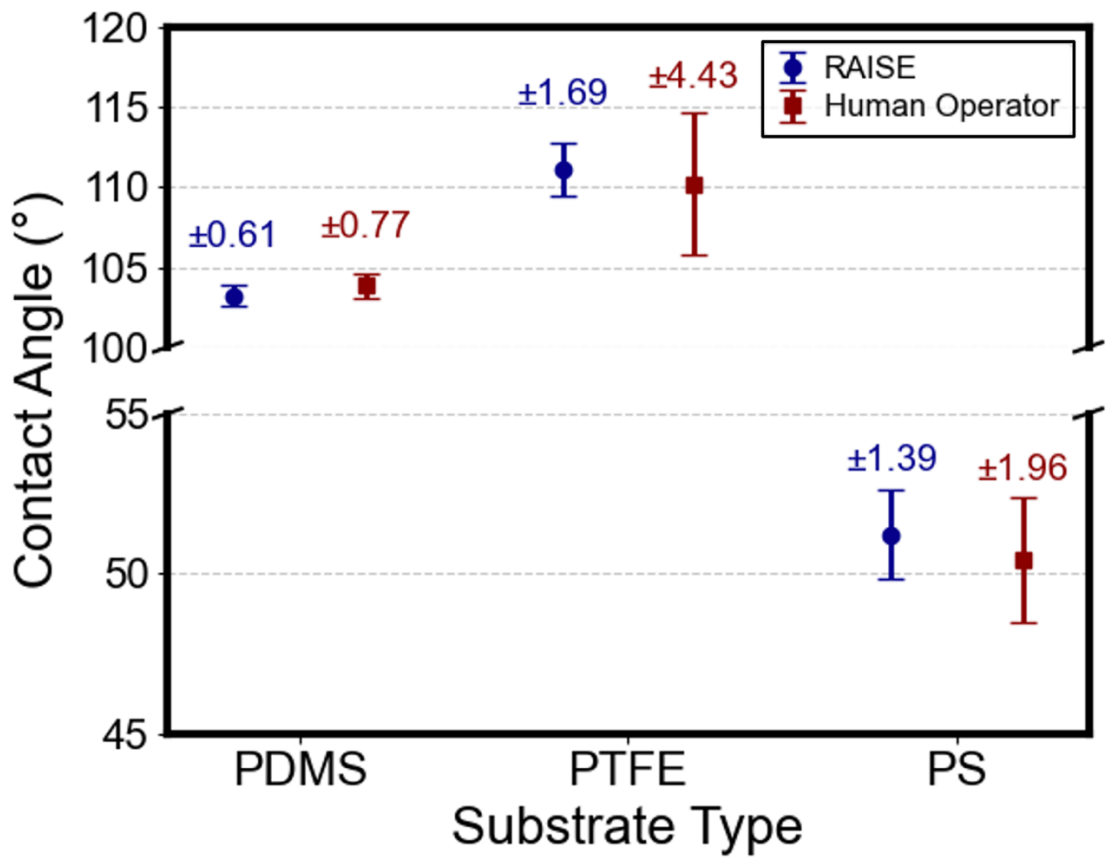

Figure 5. Static water contact angles measured by RAISE and a goniometer (manually) on PDMS, PTFE, and PS surfaces. RAISE shows strong agreement with reduced variability.

3.2. Closed-Loop Optimization of Water-Ethanol Mixtures

To evaluate the performance of our autonomous experimental platform, we conducted a fully closed-loop optimization campaign on water-ethanol mixtures using single-objective BO. This marks the first complete deployment of our system in which all steps, including experiment selection, liquid dispensing, droplet imaging, contact angle measurement, and decision-making, were executed autonomously. The goal was to identify formulations that achieve a contact angle within a target range of 85-90°, which is representative of moderately non-wetting surfaces relevant to various soft material applications.

Over 101 iterations (each representing a unique formulation tested within the predefined space) the BO directed the search within a predefined formulation space ranging 0-50 vol% ethanol, sampled at 0.5% intervals. Figure 6 summarizes the experimental outcomes. In Figure 6 (a), we observe the typical monotonic decrease in contact angle with increasing ethanol concentration.

Importantly, the red data points represent the experiments that fell within the desired contact angle range. These target-hitting formulations primarily appeared in the 25-40% ethanol region.

Figure 6 (b) presents contact angle measurements for different unique formulations. The distribution indicates that BO initially explored a broad range of conditions but rapidly optimized for conditions yielding near-target responses. The clustering of red points in the middle of the iteration range reflects the capacity of BO to exploit what it has learned about the system to repeatedly sample high-value regions. This behavior contrasts with brute-force or grid-based screening, which highlight the efficiency of BO in reducing the number of required experiments. Figure 6 (c) shows the corresponding ethanol concentrations over the iteration numbers in the campaign. This shows that BO can quickly find formulations that fall within the target contact angle range. The ability of the system to adjust its sampling trajectory in real time, guided by experimental feedback, highlights the value of closed-loop, data-driven experimentation. This campaign demonstrates the successful integration of single-objective BO with automated hardware for formulation discovery. The system efficiently navigated the chemical space, reproduced known trends in ethanol-water wetting behavior [53,54], and achieved target performance within the first 4% of experiments. This water-ethanol campaign was the first test run of RAISE, showing that the system can be extended to multi-component formulations, as detailed in the following sections.

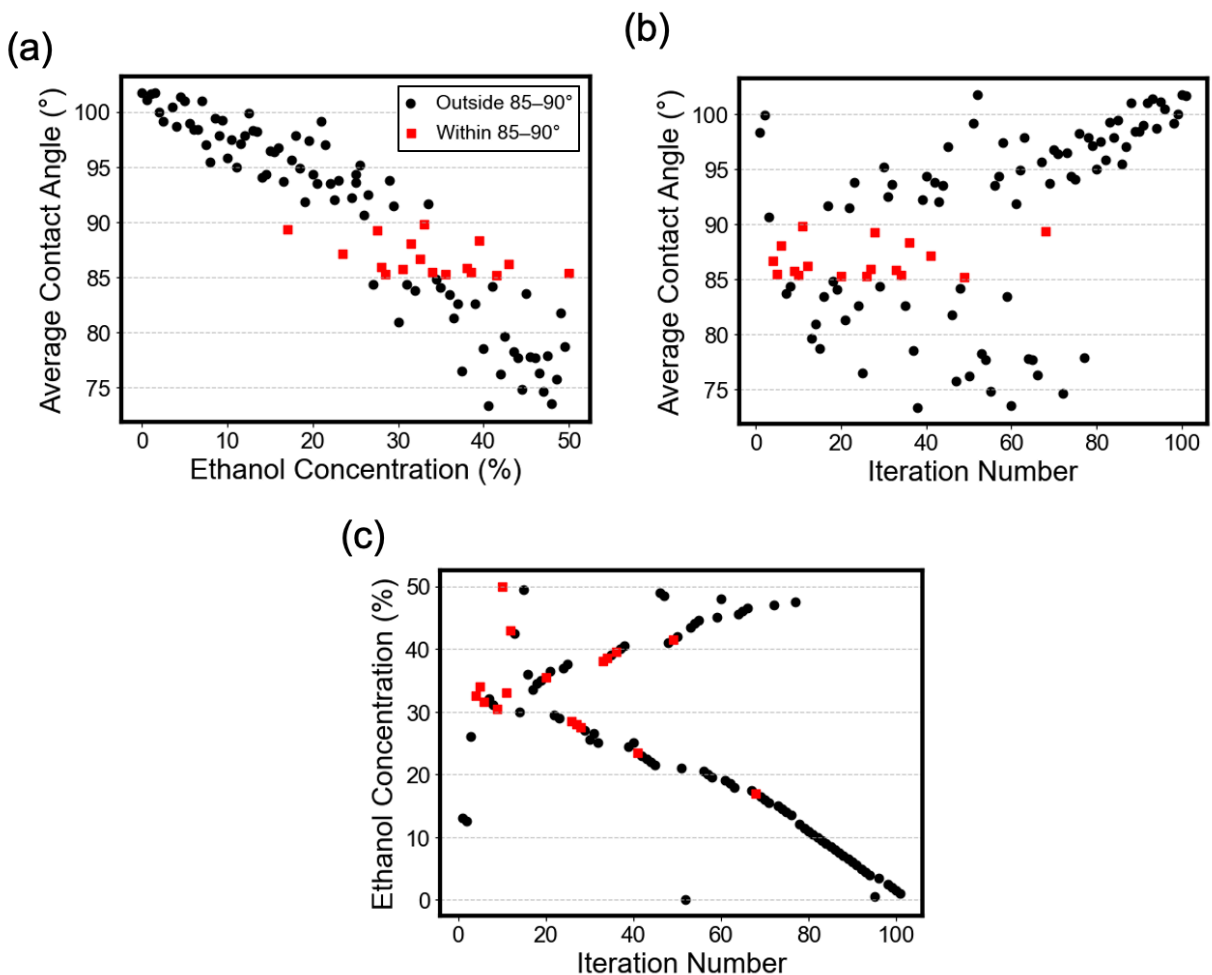

Figure 6. Fully closed-loop optimization of water-ethanol mixtures using BO. All 101 experiments were autonomously selected and executed by the BO-driven platform. Red squares indicate formulations that achieved the target contact angle range of 85-90°. Average contact angle (means of replicate measurements) as a function of (a) ethanol concentration and (b) iteration number. (c) Ethanol concentration over iteration number, showing BO's adaptive exploration of the formulation space.

3.3. Surfactant Performance and Ethanol-Induced Plateau Shifts in Wettability

To establish a baseline understanding of how individual surfactants influence wettability, we performed a systematic screening of six surfactants in water across a concentration range of 0-0.6% (w/v), in 0.04% intervals, without any optimization algorithm. As shown in Figure 7 (a), all formulations were evaluated for their ability to reduce the contact angle on the target substrate. Each surfactant demonstrated a distinct trend, indicative of its molecular characteristics and surface activity. A sharp decrease in contact angle was observed with even small additions of surfactant compared to pure water, consistent with the typical behavior of surfactant adsorption at interfaces [55]. DGO and Titrin X-100 exhibited the most pronounced reduction in contact angle, maintaining values consistently below 55° for all tested concentrations. This suggests a strong affinity for the interface and rapid surface saturation. In contrast, SDS 99% and SDS 94% exhibited distinct wettability profiles across the concentration range. This difference is likely attributable to their purity levels, suggesting that even minor variations in purity can impact adsorption behavior and surface tension reduction. SDS 99% showed a gradual decrease in contact angle with increasing concentration, plateauing around 70°, consistent with the onset of its critical micelle concentration (CMC), which limited further reduction in static contact angle. The SDS 94% sample has approximately 5.5% non-active components, including inorganic salts (NaCl and $Na_2SO_4$), unsulfated matter, and moisture. Prior works have shown that adding salts such as NaCl or $Na_2SO_4$ to anionic surfactant solutions can either enhance or hinder wetting depending on concentration, due to changes in CMC and electrostatic interactions [56,57]. Interestingly, low- and high-purity forms of TWEEN20 yielded nearly identical contact angle profiles, indicating that, in this case, purity had minimal influence on wetting behavior at the tested scales.

Exploring different formulation combinations with surfactants highlights the role of ethanol in formulations-wettability relationship. Ethanol, due to its low surface tension and lower contact angle, can help reduce the plateau typically observed with surfactants. For example, in Figure 7 (a), SDS 99% alone reaches a plateau in static contact angle around 72°; increasing its concentration does not significantly reduce the contact angle further. In comparison, Figure 7 (b) shows that the contact angle decreases with increasing ethanol content, which reaches values below 50° when using 80% ethanol. This effectively lowers the plateau. As expected, higher ethanol concentrations also eliminate the initial slope of the curves and flatten them, since ethanol exerts a stronger influence on wettability than SDS 99% in this experiment.

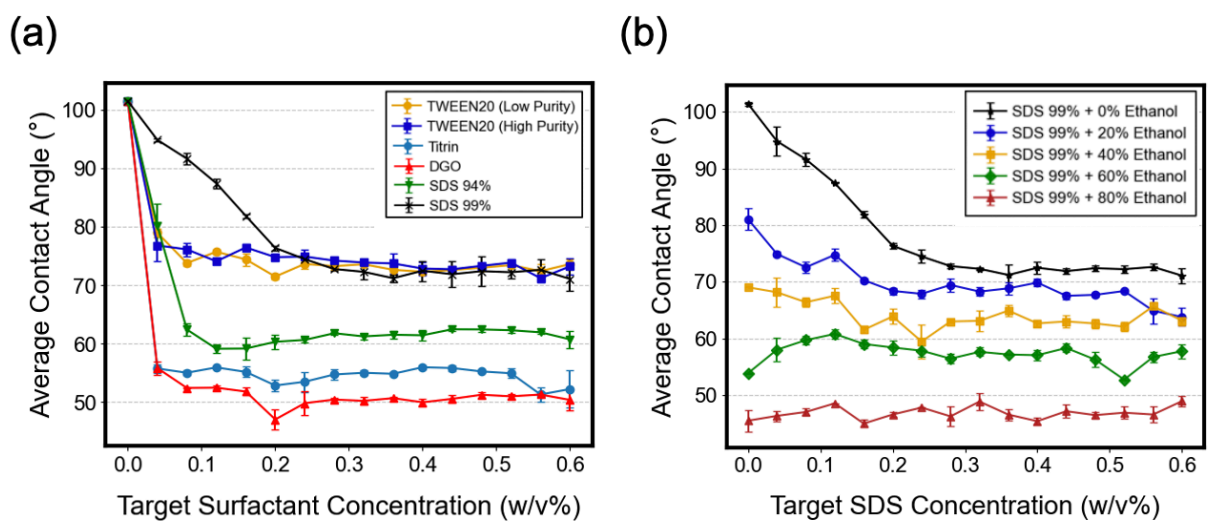

Figure 7. (a) Contact angle trends for various surfactants with increasing concentration. (b) Varying ethanol concentrations shift the SDS 99% contact angle plateau to lower surfactant concentrations.

3.5. Improving Wettability and Cost Efficiency Through the RAISE Platform

In practical applications, inconsistencies in material quality and cost limitations often make it challenging to achieve the desired wettability. Our RAISE platform provides a way to quickly screen and optimize surfactant formulations to address these issues. In this section, we show two examples where RAISE helps improve both performance and cost efficiency: (I) Formulation optimization using multi-objective Bayesian optimization (2) Compensating for surfactant purity.

Figure 73.5.1. Targeted Formulation Optimization with Multi-Objective Bayesian Optimization

To demonstrate how a BO campaign can effectively seek out the design of multi-component formulations that optimize multiple parameters, a closed-loop BO campaign was designed to test mixtures of low-purity SDS (94%) and Tween 20. The goal was to identify formulations that replicate the wettability performance of high-purity SDS (99%) while minimizing total surfactant usage. This goal was informed by the concentration dependent wettability profile shown in **Error! Reference source not found.** (a). The initial set of formulations tested by the BO client was selected randomly to ensure unbiased exploration of the formulation space at the start of the campaign.

A two-objective BO campaign was designed to optimize target parameters (i.e. static contact angle and total surfactant concentration) using a desirability objective. In the campaign, a desirability score is calculated for each SDS 94%-Tween 20 formulation that the BO client recommends for testing. The desirability score is calculated by taking a weighted geometric mean of each normalized target parameter. Each target is first normalized to fall within the range [0,1]. After normalization, every target is assigned a weight that reflects its relative importance in the optimization problem. Targets with higher weights have greater influence over the score. To combine multiple target objective values into a single score, the weighted geometric mean of each normalized target parameter value is calculated, which has the advantage of penalizing poor performance in any one target more strongly than a simple arithmetic mean. The overall desirability score $D$ is defined as shown in Equation (2).

$$D = \left(\prod_{i=1}^{n} t_i^{w_i}\right)^{1/\sum_{i=1}^{n} w_i} \tag{2}$$

where $t_i \in [0,1]$ is the normalized value of the target, $w_i$ is the weight assigned to that target, and $n$ is the total number of targets considered. In this study, the weight ratio between minimizing surfactant usage and achieving the target contact angle was set to [1:1], reflecting equal priority given to both objectives in the optimization process. The effects of altering this weighting ratio were explored in additional simulations (see Figure S1, Supporting Information), which demonstrates that heavier emphasis on either objective resulted to trade-offs in optimal formulation discovery. The aggregation of desirability scores for each tested formulation determines which formulations the BO client recommends in the future. In total, the BO client directed the experiment orchestrator to test 135 different combinations of SDS 94% and Tween 20 ranging from concentrations of 0.04-0.60% and 0.04-1.2%, respectively. To guide the exploration toward physically realistic and relevant regions, the BO client was set to search within a contact angle range of ±50° from the target of 72°, covering a range from 22° to 122°. To evaluate the impact of this constraint, we simulated campaigns using a range of alternative contact angle bounds (see Figure S2, Supporting Information). Results showed that tighter constraints led to fewer optimal formulations being identified and often delayed the overall material discovery.

To identify top-performing candidates, we defined a set of optimal formulations that met two conditions: (i) a desirability score greater than or equal to 0.90, and (ii) a static contact angle within two standard deviations of the target value of 72°, corresponding to the wettability plateau observed for SDS 99%. The standard deviation was calculated to be 0.805° based on replicate measurements of water droplets on the RAISE stage (Table S2, Supporting Information), yielding an acceptable wettability range of 70.39° to 73.61°. This range is also visualized as a shaded region in Figure 8 (d). It is important to note that while the desirability score is designed to reward formulations that are closer to the target contact angle and use less total surfactant, not all high-desirability formulations are classified as optimal formulations. To accurately identify formulations that satisfied the wettability condition, a binary mask was applied to identify only those high-scoring formulations that fall within the defined contact angle range. Among these, the six formulations with the highest desirability scores were designated as optimal formulations, which are listed in Table 1. The most optimal formulations center around formulations with 0.1% SDS and 0.16% Tween 20.

Table 1. Top-performing SDS 94%-Tween 20 formulations identified by the two-target desirability optimization campaign. These formulations achieved static contact angles within the target range while minimizing total surfactant content (desirability ≥ 0.9).

| SDS 94% | TWEEN20 | Static Contact Angle | Desirability | Two-Target Objective Experiment Number | Single-Target Objective Experiment Number |
|---|---|---|---|---|---|
| 0.08 | 0.20 | 71.14 | 0.92 | 20 | 80 |
| 0.08 | 0.24 | 73.25 | 0.91 | 23 | 79 |
| 0.04 | 0.12 | 73.26 | 0.95 | 25 | 99 |
| 0.12 | 0.12 | 73.55 | 0.92 | 34 | 75 |
| 0.16 | 0.12 | 71.15 | 0.92 | 37 | 70 |
| 0.16 | 0.16 | 72.01 | 0.91 | 38 | 36 |

To evaluate the performance of the multi-objective optimization campaign compared to a single-objective campaign, a single-objective BO campaign was simulated by feeding in the experimental data collected from the closed-loop self-driving RAISE platform. Whereas the multi-objective campaign sought to match a target contact angle value and minimize total surfactant usage, the single-objective campaign sought to only match a target contact angle value. As shown in **Error! Reference source not found.** (b), the two-objective campaign identified six optimal formulations between experiments 20 and 38, representing 14.8%–28.1% of the total campaign. In contrast, the same formulations were identified during the single-objective campaign (**Error! Reference source not found.** (c)) between experiments 36 and 99, representing 26.7%–73.3% of the campaign. Furthermore, the two-objective strategy identified the optimal formulations within a narrower experimental window, indicating a more efficient and targeted search. These results demonstrate that integrating multiple objectives into the optimization process enables more efficient discovery of liquid formulations that satisfy multiple design constraints. This trend was consistently observed across multiple simulated campaign runs with randomly selected initial formulation tested (Figure S3, Supporting Information), which further confirms the robustness and efficiency advantage of the multi-objective optimization.

To further visualize the relationship between formulation composition, wettability, and optimization outcomes, we generated a scatter plot of all tested formulations with contact angle on the y-axis and total surfactant concentration on the x-axis (Figure 8 (d)). Each point is colored based on its desirability score, and optimal formulations are highlighted in red stars. The shaded region centered at 72° represents the acceptable static contact angle range (70.39°–73.61°), defined as within two standard deviations from the target. The optimal formulations cluster within this region and are positioned toward the lower end of the surfactant concentration axis, indicating that they simultaneously meet the wettability requirement while using less surfactant. In contrast, several formulations with high desirability scores greater than 0.90 (shown in blue) are not classified as optimal formulations because they fall outside the target contact angle range or require higher total surfactant. This highlights that desirability alone is not sufficient, and that additional constraints, such as restrictions in key target parameters within acceptable values, are necessary to fully capture the potential of the RAISE system and find formulations that meet user-defined criteria. To further contextualize these observations, we generated contour plots of key target properties across the formulation search space (Figures S4, Supporting Information).

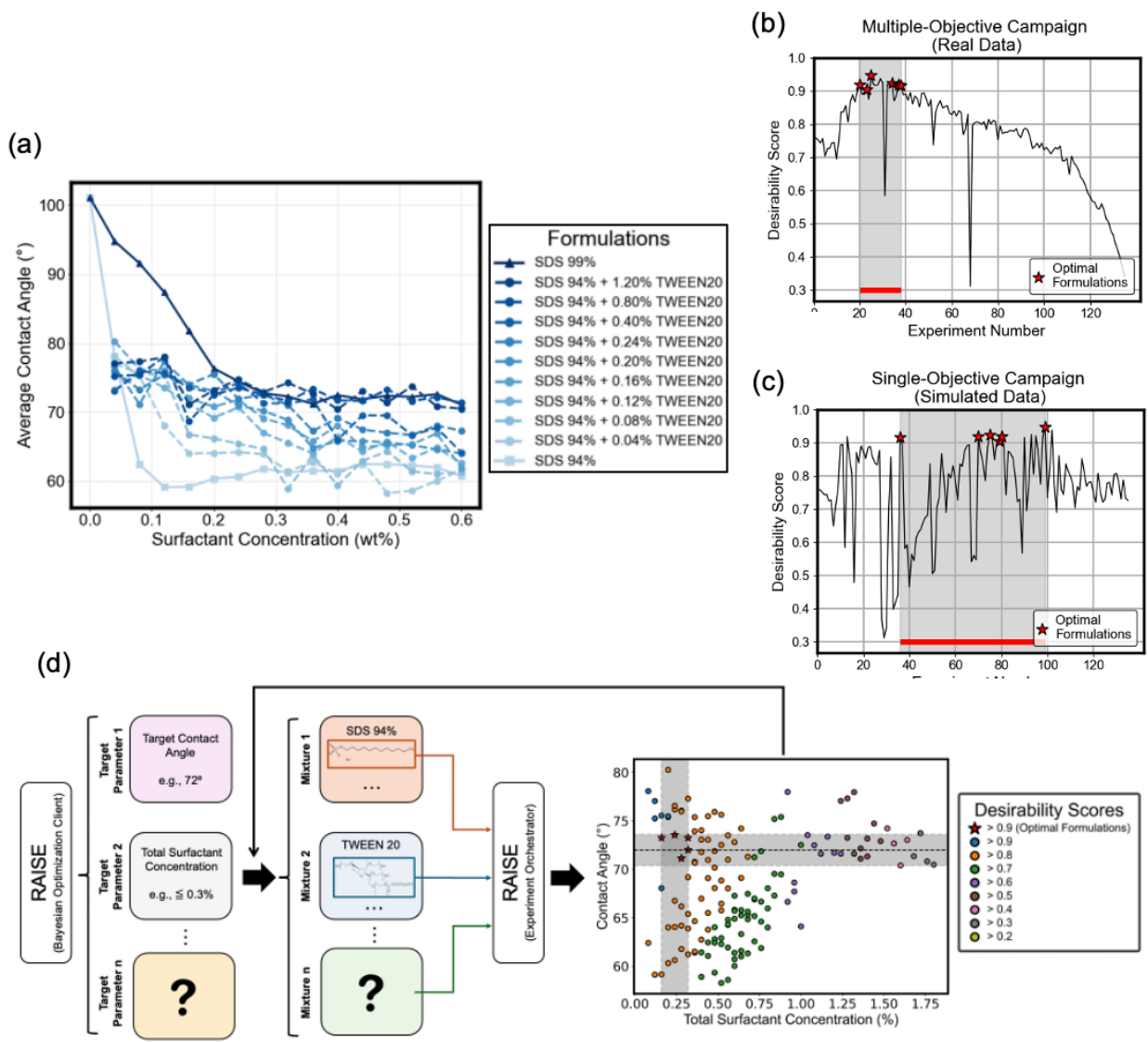

Figure 8. a) Adding low-cost Tween 20 to SDS 94% recovers wettability, matching high-purity SDS (99%) performance. Desirability scores from (b) multi-objective and (c) single-objective BO campaigns. Optimal formulations (red star) were found earlier and more. (d) RAISE workflow and outcomes. Scatter plot shows contact angle vs surfactant concentration, colored by desirability score. Red stars mark optimal formulations within the target contact angle range (shaded in gray) and desirability ≥ 0.9.

3.5.2. Compensating for Surfactant Purity Using Automated Screening

As shown in Figure 7 (a), the wettability of individual surfactants in solution plateau as the surfactant concentration increases and that the combination of multiple chemical reagents in a solution can alter the total wettability of the system. Individual surfactants exhibit concentration-dependent wettability plateaus, suggesting multiple formulation can achieve equivalent performance while enabling optimization of secondary objectives such as cost. We exploited this principle to address purity-related variability between SDS 99% and SDS 94%, where the lower purity grade showed substantially higher contact angles (Figure 7a). RAISE identified that blending low-cost Tween 20 with SDS 94% could compensate for the reduced purity. Among the most optimal formulations found by multi-objective BO campaign, the combination of 0.04wt% SDS 94% and 0.12wt% Tween 20 was found to use the least amount of total surfactant (Table 2). In comparison, 0.28wt% is the minimum concentration of SDS 99% to achieve similar wettability. This substitution has a potential of economic advantage: SDS 99% costs \$3.38/g (Sigma-Aldrich) vs \$0.0103/g for SDS 94%. A l Liter solution using the compensated formulation costs \$0.0196 compared to \$9.56 for equivalent SDS 99% performance but with a 490X cost reduction. While surfactant blending for performance compensation has been well established in the literature [58–60], this BO-guided approach has the potential to enable rapid identification of synergistic components that might be overlooked in manual workflows, especially when balancing multiple competing objectives.

## 4. Conclusion

In summary, RAISE represents the first fully autonomous self-driving laboratory integrating liquid formulation optimization, contact angle measurement, and BO in a close-loop workflow. Benchmarking against conventional goniometry demonstrated equivalent accuracy with improved precision showing 20% - 62% reduction in standard deviation, and 10X higher throughout. The platform successfully navigated formulation spaces through autonomous experimentation, identifying water-ethanol mixture meeting target wettability within 100 iterations. By automating formulation characterization workflows, RAISE eliminates conventional trial-and-error bottlenecks while enabling systematic exploration of higher dimensional chemical spaces. The modular hardware design and flexible optimization framework can extend beyond surfactant screening to applications in coating, biomedical devices, and personal care products where surface wettability governs performance.

RAISE: A self-driving laboratory for interfacial property formulation discovery

Mohammad Nazeri [1†], Sheldon Mei [2†], Jeffrey Watchorn [3†], Alex Zhang [4], Erin Ng [2], Tao Wen [4], Abhijoy Mandal [5], Kevin Golovin [4], Alán Aspuru-Guzik [3,5,6,7], Frank Gu [1,2,3*]

Affiliations
[1] Institute of Biomedical Engineering, University of Toronto, 164 College St, Toronto, ON M5S 3G9
[2] Department of Chemical Engineering and Applied Chemistry, University of Toronto, 200 College St, Toronto, ON M5S 3E5
[3] Acceleration Consortium, 700 University Avenue, Toronto, ON M5G 1X6
[4] Department of Mechanical & Industrial Engineering, University of Toronto, 5 King's College Rd, Toronto, ON M5S 3G8
[5] Department of Computer Science, University of Toronto, Toronto, Ontario, M5S 3H6, Canada
[6] Department of Chemistry, University of Toronto, Toronto, Ontario, M5S 3H6, Canada
[7] Vector Institute for Artificial Intelligence, 661 University Ave. Suite 710, ON M5G 1M1, 15 Toronto, Canada

Corresponding author's email address
f.gu@utoronto.ca

To assess how consistent the contact angle measurements were, the standard deviations from both RAISE and the manual goniometer were compared across three different surfaces. An F-test was used to determine if one method was actually more reliable than the other, or if any differences observed were just random (Table S1). The F-test compares the variances (the standard deviations (SD) squared) between the two methods, and the p-values indicate how likely it is that these differences would occur just by chance. RAISE showed more consistent results than the manual goniometer across all surfaces tested. This was especially evident with PTFE, where RAISE's measurements were significantly more reliable (p-value = 0.0106). This improvement occurs because RAISE automates the droplet placement and image capture, which eliminates much of the human error that can occur during manual operation.

Table S1. Comparison of measurement variability between RAISE and manual goniometer.

| Substrate | RAISE SD (°) | Goniometer SD (°) | F-statistic | p-value | % Reduction in SD |
|---|---|---|---|---|---|
| PDMS | 0.61 | 0.77 | 1.593 | 0.2243 | 20.8 |
| PS | 1.39 | 1.96 | 1.988 | 0.2155 | 29.1 |
| PTFE | 1.69 | 4.43 | 6.781 | 0.0106 | 61.9 |

To evaluate the consistency and stability of our contact angle measurement system, 3 µL water droplets were deposited onto PDMS substrates positioned at five different stage levels (A to E). Each level was tested across nine sample positions to account for variations across a single substrate. The measured contact angles (°) are reported in Table S2. This assessment was conducted to estimate the system's standard deviation (0.805) and ensure reliability across different measurement zones. The average contact angles observed in this dataset are slightly lower than those shown in the benchmarking results (Figure 5 of the main manuscript). This minor discrepancy is mainly due to the gradual degradation of this specific PDMS surface over time, which affects its hydrophobicity.

Table S2. Contact angle (°C) of 3 µL water droplets on PDMS at five stage levels. Standard deviation: 0.805°.

|  | Contact Angle (°C) | | | | | | | | |
|---|---|---|---|---|---|---|---|---|---|
| Stage Level | Sample 1 | Sample 2 | Sample 3 | Sample 4 | Sample 5 | Sample 6 | Sample 7 | Sample 8 | Sample 9 |
| A | 100.947 | 100.211 | 100.876 | 100.601 | 99.193 | 100.229 | 99.868 | 98.127 | 99.54 |
| B | 99.918 | 99.096 | 99.489 | 99.25 | 100.196 | 100.275 | 99.886 | 99.87 | 100.277 |
| C | 99.902 | 99.474 | 99.423 | 99.643 | 100.016 | 99.951 | 100.244 | 99.624 | 100.146 |
| D | 101.359 | 100.421 | 100.146 | 100.275 | 100.194 | 100.407 | 100.713 | 100.292 | 100.277 |
| E | 101.995 | 101.77 | 101.601 | 100.485 | 100.997 | 101.502 | 100.887 | 101.341 | 102.034 |
| Total Average | 100.288 | | | | | | | | |
| Total Standard Deviation | 0.805 | | | | | | | | |

We conducted simulations with varying weight ratios between matching the contact angle and minimizing total surfactant usage to investigate the effects of prioritizing different design goals on optimization results. Testing whether prioritizing one goal over the other would affect the algorithm's capacity to identify the optimal formulations was the aim. The weights are presented in the format [target contact angle, minimal surfactant usage]. As shown in Figure S1, when both objectives are equally weighted ([1, 1]), the algorithm efficiently discovers formulations that balance performance and surfactant reduction. Increasing the weight on surfactant minimization (e.g., [1, 3] or [1, 5]) leads to leaner formulations but may sacrifice some accuracy in contact angle.

On the other hand, more accurate wettability is achieved by giving priority to contact angle matching (e.g., [3, 1] or [5, 1]), but with a higher surfactant usage.

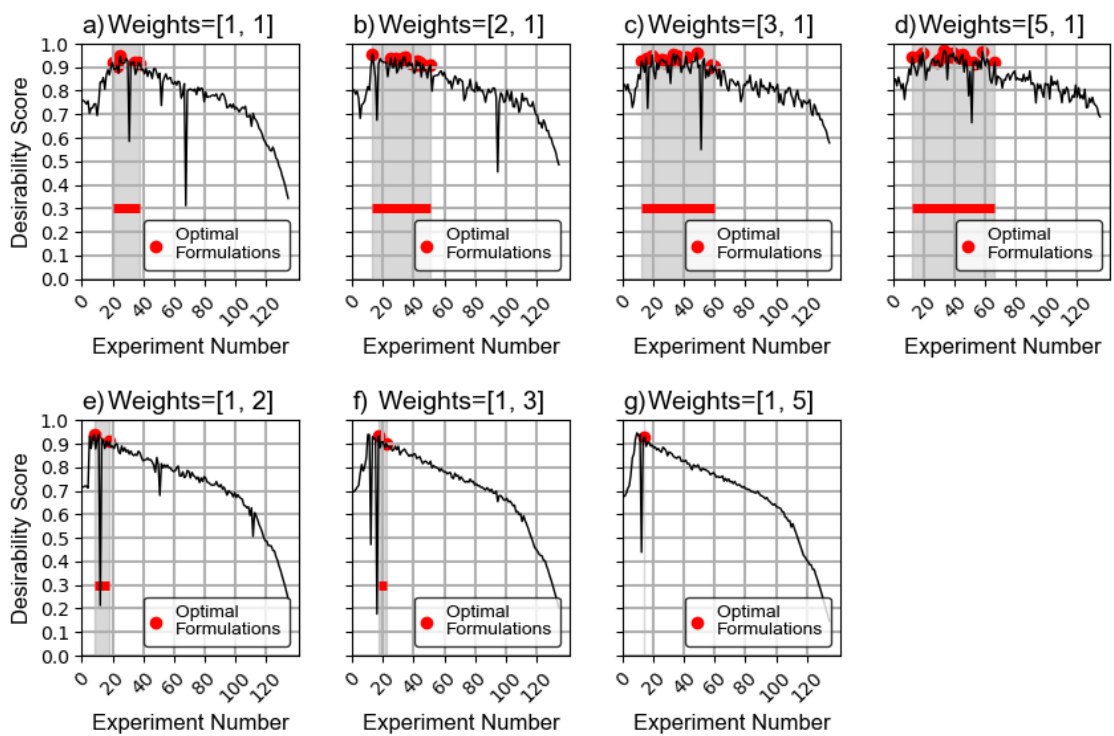

Figure S1. Impact of different weighting ratios between contact angle matching and surfactant minimization on BO performance. Giving equal ratio to both goals lead to balanced results, while shifting priorities alters the composition of optimal formulations, and the time necessary to identify them.

To understand how a strict contact angle exploration range affects the optimization process, we performed simulations using different exploration ranges around the target contact angle of 72°. These bounds ranged from very wide (±50°) to very narrow (±0.805°), which corresponds to two standard deviations based on experimental measurements. The goal was to explore how the precision of the design exploration range influences the ability of the optimization algorithm to find desirable formulations. As shown in the Figure S2, when the contact angle constraint is loose, the algorithm identifies high-desirability formulations more frequently and earlier in the campaign. In contrast, tighter bounds make it harder to meet the contact angle requirement, resulting in fewer qualifying formulations and longer discovery times.

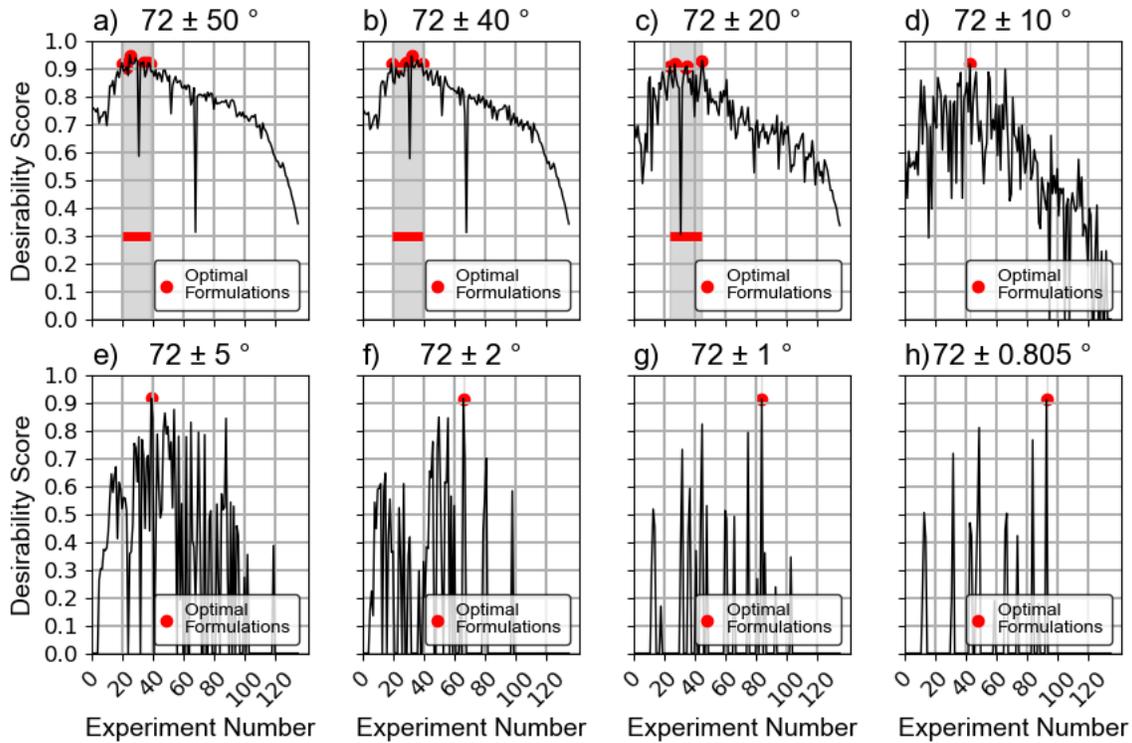

Figure S2. Effect of changing contact angle bounds on multi-objective BO performance. Stricter limits make it harder to find good formulations quickly, while wider ranges help the algorithm find solutions faster.

To investigate the effect of initial formulation selection, a simulated test was performed by varying the random starting formulation for both single- and multi-objective optimization. As show in Figure S3, each line shows how the desirability score changed over time for a different random starting point. The results demonstrate that multi-objective optimization was able to find optimal formulations (those with high desirability that also fall within the target contact angle rang) earlier and more consistently across all tests. This suggests that the multi-objective method is more efficient and less sensitive to how the experiment begins, making it more reliable in practice.

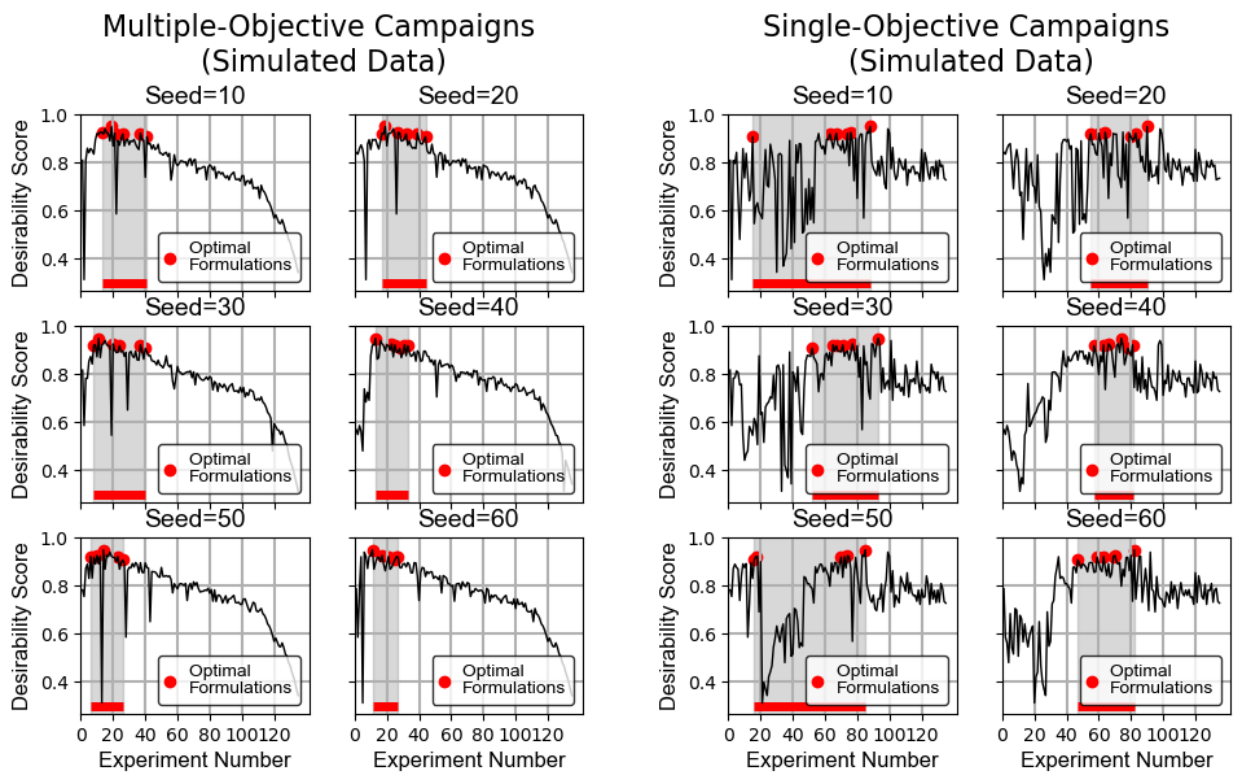

Figure S3. Comparison of multi-objective and single-objective optimization using different random starting formulations. Multi-objective optimization finds optimal formulations earlier and more consistently, no matter where it starts. Random seeds were set at the beginning of experiments to initialize the random number generator and to ensure reproducibility in the order of recommended experiments.

To visualize the formulation landscape explored by the BO campaign, we generated contour plots showing the variation of key target metrics–static contact angle, total surfactant concentration, and overall desirability–across the tested formulation space (Figure S4). Each plot maps SDS 94% and Tween 20 concentrations on the vertical and horizontal axes, respectively. Figure S4 (a) shows how the contact angle varies across the formulation space, with a broad plateau region around 72°, matching the desired wettability profile of high-purity SDS (99%). Figure S4 (b) illustrates the total surfactant concentration, which increases linearly with the combined component inputs, and serves as a penalty metric in the desirability function. Figure S4 (c) presents the computed desirability score, which combines both target parameters into a single optimization objective. Notably, the optimal formulations (highlighted in red) cluster within a region of high desirability, low surfactant content, and contact angles near the target.

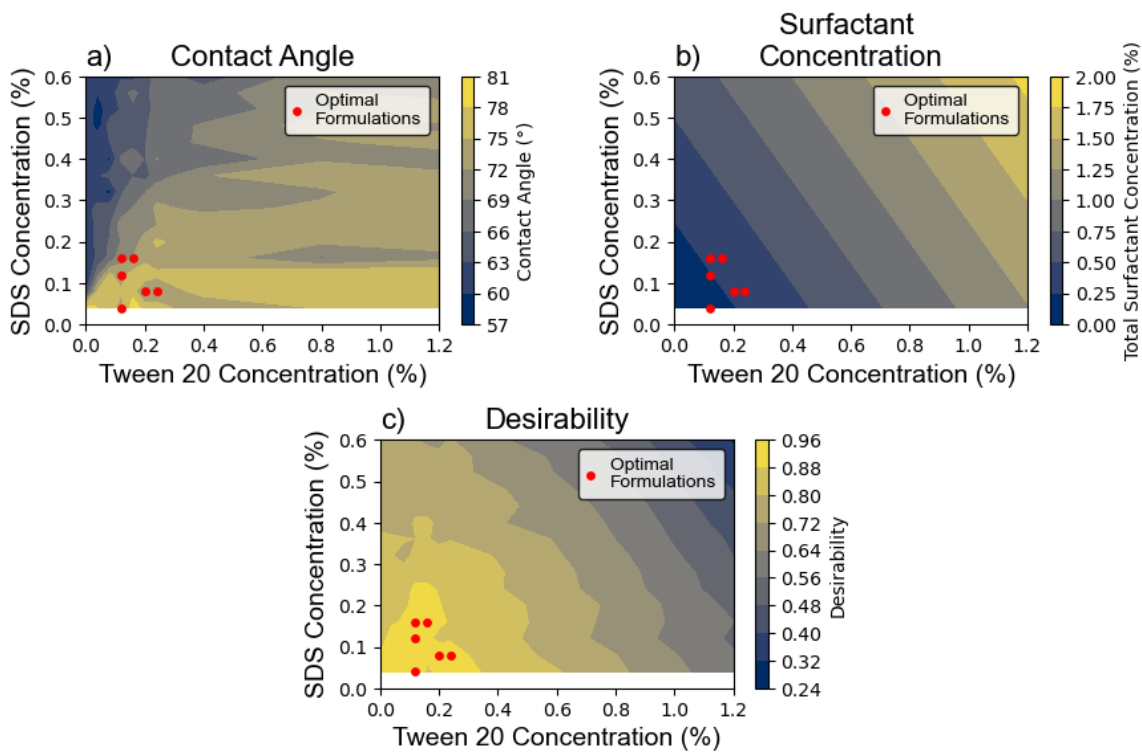

Figure S4. Contour plots of contact angle (a), total surfactant concentration (b), and desirability score (c) across the SDS 94%-Tween 20 formulation space. Red dots indicate Optimal formulations.